\documentclass[a4paper,fleqn]{cas-dc}

\usepackage{xcolor}
\usepackage[authoryear]{natbib}
\usepackage{hyperref}
\usepackage{graphicx}
\usepackage{subfigure}
\usepackage{epstopdf}
\usepackage{color}
\usepackage[noend]{algpseudocode}
\usepackage{algorithmicx,algorithm}

\usepackage{amsmath}
\usepackage{amsfonts}
\usepackage{algorithm}
\usepackage{algpseudocode}
\usepackage{graphicx}
\usepackage{subcaption}
\usepackage{amsmath}
\usepackage{enumitem}
\usepackage{multirow}
\usepackage{diagbox}
\usepackage{booktabs}

\newtheorem{myexample}
{Example}

\def\tsc#1{\csdef{#1}{\textsc{\lowercase{#1}}\xspace}}
\tsc{WGM}
\tsc{QE}
\tsc{EP}
\tsc{PMS}
\tsc{BEC}
\tsc{DE}

\begin{document}
\begin{sloppypar}
	\let\WriteBookmarks\relax
	\def\floatpagepagefraction{1}
	\def\textpagefraction{.001}
	\let\printorcid\relax
	\shorttitle{}
	\shortauthors{R. Miao et~al.} 

	\title [mode = title]{Using Experience Classification for Training Non-Markovian Tasks}



	\author[1]{\textcolor[RGB]{0,0,1}{Ruixuan Miao}}
	\ead{22031212106@stu.xidian.edu.cn}

	\author[1]{\textcolor[RGB]{0,0,1}{Xu Lu}}
	\address[1]{Institute of Computing Theory and Technology and State Key Laboratory of Integrated Services Networks, Xidian University, PR China}
	\cormark[1]
	\ead{xlu@xidian.edu.cn}

	\author[1]{\textcolor[RGB]{0,0,1}{Cong Tian}}
	\ead{ctian@mail.xidian.edu.cn}

    \author[1]{\textcolor[RGB]{0,0,1}{Bin Yu}}
	\ead{byu@xidian.edu.cn}

    \author[1]{\textcolor[RGB]{0,0,1}{Zhenhua Duan}}
	\ead{zhhduan@mail.xidian.edu.cn}

	\cortext[cor1]{Corresponding author.} 

	\fntext[fn1]{Equal Contribution.}


	\begin{abstract}
		Unlike the standard Reinforcement Learning (RL) model, many real-world tasks are non-Markovian, whose rewards are predicated on state history rather than solely on the current state. Solving a non-Markovian task, frequently applied in practical applications such as autonomous driving, financial trading, and medical diagnosis, can be quite challenging. We propose a novel RL approach to achieve non-Markovian rewards expressed in temporal logic LTL$_f$ (Linear Temporal Logic over Finite Traces). To this end, an encoding of linear complexity from LTL$_f$ into MDPs (Markov Decision Processes) is introduced to take advantage of advanced RL algorithms. Then, a prioritized experience replay technique based on the automata structure (semantics equivalent to LTL$_f$ specification) is utilized to improve the training process. We empirically evaluate several benchmark problems augmented with non-Markovian tasks to demonstrate the feasibility and effectiveness of our approach.
	\end{abstract}


		

	\maketitle

	\section{Introduction}

    Reinforcement Learning (RL) is a well-known paradigm for autonomous decision-making tasks in complex and unknown environments \cite{sutton2018reinforcement}. It has been used to help agents learn action policies from trial-and-error experiences towards maximizing long-term rewards. RL, especially deep RL, has achieved impressive results over the past few years. The standard RL environment model is that of Markov Decision Processes (MDPs). In MDPs, rewards only depend on the current state. However, many real-world tasks are non-Markovian, where rewards are determined by a sequence of past states rather than the current state. For instance, a reward is given for bringing coffee only if one was requested earlier and has not yet to be served, which is a typical non-Markovian task. An RL agent that attempts to learn such tasks without realizing that the tasks are non-Markovian will display sub-optimal behaviors. 

 Most current research focuses on defining non-Markovian tasks by temporal logic. Camacho et al. present a means of specifying non-Markovian reward, expressed in Linear Temporal Logic (LTL) interpreted over finite traces, referred to as LTL$_f$ \cite{camacho2018non}. Non-Markovian reward functions can be translated to automata representations, and reshaped automata-based rewards are exploited by off-the-shelf MDP planners to guide search. This approach is quite limited and cannot be applied to continuous domains. A more general form for specifying non-Markovian tasks is proposed in \cite{icarte2022reward,camacho2019ltl}, called reward machines which is a kind of finite state automaton. Different methodologies, such as Q-learning and hierarchical RL, to exploit such structures are presented, including automated reward shaping, task decomposition, and counterfactual reasoning for data augmentation. The authors find a primary drawback that reward shaping does not help in continuous domains. Analogous to \cite{camacho2018non}, our previous work \cite{miao2022approach} also proposes a translation method for non-Markovian reward but upgrades it to obtain more compact policies.  

Hasanbeig et al. investigate a deep RL method for policy synthesis in continuous-state/action unknown environments under requirements expressed in LTL \cite{hasanbeig2020deep}. An LTL specification is converted to a Limit Deterministic B{\"{u}}chi Automaton (LDBA) and synchronized on-the-fly with the agent/environment. A modular Deep Deterministic Policy Gradient (DDPG) architecture is proposed to generate a low-level control policy that maximizes the probability of the given LTL formula. The synchronization process automatically modularizes a complex global task into easier sub-tasks in order to deal with the sparse reward problem.
Voloshin et al. study the problem of policy optimization with LTL constraints \cite{voloshin2022policy}. A learning algorithm is derived based on a reduction from the product of MDP and LDBA (modeling the progression of LTL satisfaction) to a reachability problem. This approach enjoys a sample complexity analysis for guaranteeing both task satisfaction and cost optimality since the authors make two strong assumptions: the availability of a generative model of the environment and a lower bound on the transition probabilities in the
underlying MDPs. However, this approach is still not extended to continuous state and action spaces.
Bozkurt et al. present an RL framework to synthesize a control policy from a product of LDBA (LTL specification) and MDPs \cite{bozkurt2020control}. The primary contribution is to introduce a novel rewarding and discounting scheme based on the B{\"{u}}chi (repeated reachability) acceptance condition.
Two simple motion planning case studies solved by a basic Q-learning implementation are illustrated to show the optimization of the satisfaction probability of the B{\"{u}}chi objective.

By studying the satisfaction of temporal properties on unknown stochastic processes that have continuous state spaces, Kazemi et al. propose a sequential learning procedure based on a path-dependent reward function obtained from the positive standard form of the LTL specification and maximized via the learning procedure \cite{kazemi2020formal}. 
Oura et al. propose a method for the synthesis of a policy satisfying a control specification described by an LTL formula, which is converted to an augmented Limit-Deterministic Generalized B{\"{u}}chi Automaton (LDGBA) \cite{oura2020reinforcement}. A product of augmented LDGBA and MDP is generated, based on which a reward function is defined to relax the sparsity of rewards.
Li et al. introduce Truncated LTL (TLTL) as a specification language, together with quantitative semantics \cite{li2017reinforcement}. They propose an RL approach to learn tasks expressed as TLTL formulae and demonstrate successfully in a toast-placing task learned by a Baxter robot. 

This paper proposes an RL approach to train non-Markovian tasks specified by LTL$_f$. Our contributions are twofold. First, we introduce an effective encoding method in order to transform non-Markovian tasks into Markovian tasks at the price of a low (linear) increase of training space. Second, we propose an automata-based method to classify experiences and design a novel automatic reward shaping mechanism in RL to improve training speed and policy quality.

The remainder of the paper is organized as follows: The second section gives a review of the basic concepts and notations used in this paper. The third section illustrates our approach in detail, including non-Markovian model construction and experience classification. In the fourth section, we provide experimental evaluations on benchmark problems to show the performance of our approach. Finally, we draw some conclusions and outline directions for future work.

\section{Preliminaries}
\label{title_page}
This section briefly presents some background concepts. We first review LTL$_f$ \cite{de2013linear}, then give the definition and a typical example related to a non-Markovian task. Finally, we describe Relational Dynamic Diagram Language (RDDL) \cite{sanner2010relational} and pyRDDLGym \cite{taitler2022pyrddlgym}. The former, widely applied in the planning community, is the standard language to model MDPs, and the latter is a toolkit to convert RDDL to Gym environment.
\subsection{Linear Temporal Logic over Finite Traces}
LTL$_f$ is a variant of LTL concentrating on expressing temporal properties over the finite sequence of states. Let $AP$ be the set of atomic propositions. The syntax of LTL$_f$ is defined as follows:
 \begin{equation}
\varphi ::= p \mid \neg \varphi \mid \varphi_1\wedge \varphi_2 \mid \bigcirc \varphi \mid \varphi_1 \text{U} \varphi_2\nonumber
\end{equation}
where $p\in AP$ is an atomic proposition, $\bigcirc$(next) and $\text{U}$(until) are temporal operators. Propositional binary connectives $\vee, \rightarrow, \leftrightarrow$ and boolean values $\top$, $\bot$ can be derived in terms of basic operators. Other temporal operators can also be derived. For example, $\Diamond$(eventually), $\Box$(always), and $\bigodot $(weak-next) are defined by:
\begin{equation}
\Diamond \varphi\equiv \top \text{U} \varphi \quad \Box\varphi\equiv\neg\Diamond\neg\varphi \quad \mathbin{\scalebox{1.0}{$\bigodot$}}\varphi\equiv Last\vee \bigcirc\varphi \nonumber 
\end{equation}
$Last$ is a specific modality of LTL$_f$ to specify properties that hold only in the final state of a trace. $\bigcirc\varphi$ denotes that $\varphi$ holds in the next state, and $\bigodot\varphi$ is similar to $\bigcirc\varphi$, but a next state is not strictly necessary.
 $\varphi_1 \text{U} \varphi_2$ indicates that $\varphi_1$ holds until $\varphi_2$ is true. $\Diamond\varphi$ means that $\varphi$ will eventually hold before the last state. $\Box\varphi$ represents that $\varphi$ holds along the whole trace. 
 
  A state $s$ is a subset of $AP$ that is true, while atomic propositions in $AP \setminus s$ are assumed to be false. LTL$_f$ formulae are interpreted over finite traces of states  $\sigma=s_0...s_n$. The semantics of LTL$_f$ is defined as follows. We say that $\sigma $ satisfies $\varphi$, written as $\sigma \models \varphi $, when $\sigma,0 \models \varphi$.
\begin{itemize}[label=\textbullet]
    \item $\sigma,i\models p$\quad \textbf{iff} \quad $p \in s_i$.
    \item $\sigma,i \models \neg \varphi$ \quad \textbf{iff} \quad $\sigma,i \not\models \varphi$.
    \item $\sigma,i\models \varphi_1 \wedge \varphi_2$ \quad \textbf{iff} \quad $\sigma,i \models \varphi_1$ and $\sigma,i\models \varphi_2$.
    \item $\sigma,i \models \bigcirc \varphi$ \quad \textbf{iff} \quad $i<n$ and  $\sigma,(i+1)\models \varphi$.
    \item $\sigma,i\models\varphi_1 \text{U} \varphi_2$ \quad \textbf{iff} \quad there exists $i\le j\le n$ such that $ \sigma,j\models\varphi_2$, and $\sigma,k \models \varphi_1$ for each $i\le k < j$.
\end {itemize}
 
 Theoretically each LTL$_f$ formula can be transformed to a Deterministic Finite Automaton (DFA) \cite{de2021compositional}. A DFA is a tuple $\mathcal{A} =\langle Q,\Sigma,\delta,q_0,F \rangle$, where $Q$ is a finite set of states, $\Sigma$ is a finite set of input alphabet, $\delta:Q\times\Sigma\to Q$ is a transition function, $q_0\in Q$ is an initial state, $F\subseteq Q$ is a set of accepting states. Let $\sigma =s_1...s_n$ be a string over the alphabet $\Sigma=2^{AP}$. $\mathcal{A} $ accepts $\sigma$ if there exists a sequence of states $q_0...q_n$ where $q_{i+1}=\delta(q_{i}, s_{i+1})$ for $0\le i < n$ and $q_n\in F$. We call the states that can never reach $F$ the error states $E \subseteq Q$. To make it more clear, we call $s$ a state and $q$ a DFA state in the sequel. 

\subsection{MDP and NMRDP}
  An MDP $M = \langle S, A, R, P, \gamma, s_{0} \rangle$ serves as a model for an agent’s sequential decision-making process. Here, $S$ is a set of states, $A$ is a set of actions, $R:S \times A\times S\to \mathbb{R}$ is a reward function, $P(s_{t+1}|s_t, a_t) \in [0,1]$ is the transition probability distribution over the set of next states, given that the agent takes action $a_t$ in state $s_t$ at step $t$ and reaches state $s_{t+1}$, $\gamma$ is the discount factor, $s_0 \in S$ is the initial state.

MDPs are limited in expressiveness due to the memoryless feature, while Non-Markovian Reward Decision Processes (NMRDPs) are more powerful by extending MDPs with non-Markovian rewards \cite{thiebaux2006decision}. An NMRDP is a tuple $NM= \langle S, A, R, P, \gamma, s_{0} \rangle$, where $S, A, P, \gamma, s_0$ are the same as those in an MDP. The only difference is the reward function $R$, which is defined as $(S\times A)^*\to \mathbb{R}$. This function indicates that non-Markovian rewards are determined by a finite sequence of states and actions. 

\begin{myexample}[\emph{Waterworld} problem]\label{ex: control knowledge}
     The Waterworld problem is a typical non-Markovian task. The environment comprises a two-dimensional container with balls of diverse colors inside. Each ball travels in one direction at a constant speed and rebounds when it hits the boundary. The agent, depicted by a white ball, can accelerate or decelerate in any direction. The goal is to touch colored balls in a specific order. 

\begin{figure}
\centering
\includegraphics[width=0.4\textwidth]{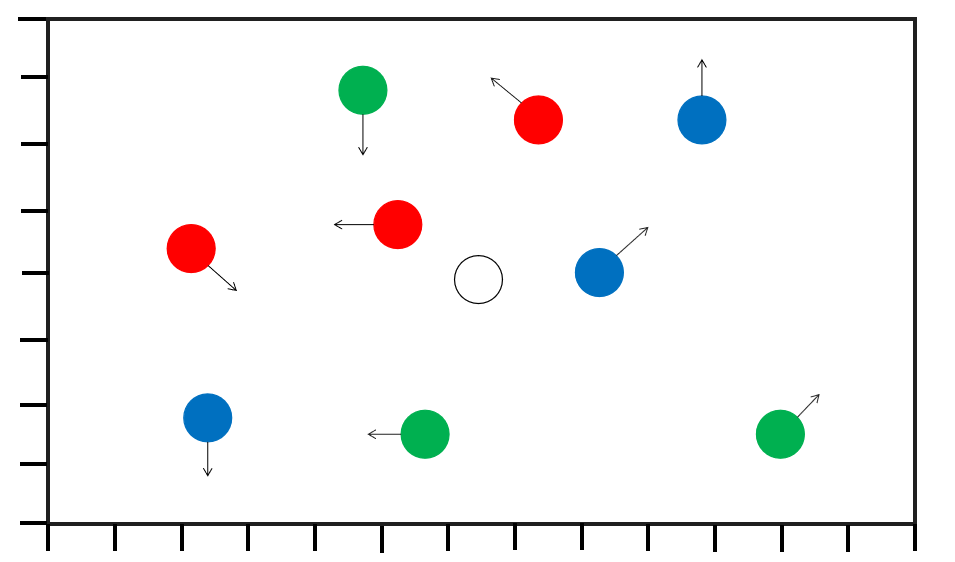}
\caption{An instance of the \emph{Waterworld} problem} 
\end{figure}
As shown in Figure 1, there are three kinds of colored balls, and the goal is first to touch a red ball and then a green one. This non-Markovian reward can be defined as the LTL$_f$ formula below, where $r$ and $g$ mean touching a red and a green ball respectively. The formula restricts that the agent cannot touch any ball except red ones at first. After that, it can only touch the green ones.
\begin{equation}
\varphi = \left ( \neg r \wedge \neg g \right ) \emph{U}((r \wedge \neg g) \wedge  \bigcirc ((\neg r \wedge  \neg g)\emph{U} g))  \nonumber
\end{equation}
\end{myexample}

\subsection{pyRDDLGym}
OpenAI Gym (Gym for short) is a toolkit for reinforcement learning research \cite{brockman2016openai}. It includes a growing collection of benchmark problems that expose a common interface, which enables people to share their results and compare the performance of algorithms. Gym provides uniform data types to define observation and action spaces, such as Box, Dict and Discrete. Convenient wrappers, utilities, and tests included in Gym also make it easy to create new environments.

RDDL is a modeling language that allows for a
lifted, compact representation of factored MDPs. In RDDL, states, actions, and observations (discrete or continuous) are parameterized fluents, and the evolution of a fully or partially observed  (stochastic) process is specified via (stochastic) functions over the next state fluents conditioned on the current state and action fluents.
A fluent is represented by predicates that can change over time. 
RDDL format is the preferred option to model the domains in the International Probabilistic Planning Competition (IPPC). Current state-of-the-art MDP planners such as Prost \cite{keller2012prost} and its variants are typically based on UCT-search techniques. Their performance suffers in large problems that require significant lookahead. Moreover, the MDP planners do not support several vital RDDL constructs, e.g., continuous action fluents and state fluents. 

To overcome the problems above, researchers provide a Python framework named pyRDDLGym for auto-generating Gym environments from the RDDL declarative description. Since RDDL describes the discrete-time step evolution of fluents using conditional probability functions, it is well compatible with the Gym step scheme. Therefore, RL algorithms can be applied to solve RDDL models with continuous data types. Currently, the pyRDDLGym toolkit is the official evaluation system for the 2023 IPC RL and planning track\footnote{https://ataitler.github.io/IPPC2023/}. Additional components and structures have been supplemented to RDDL to increase expressivity and accommodate learning interaction types. For instance, a termination block is added to end an episode explicitly.

\begin{myexample}[\emph{Waterworld} problem cont.]
RDDL mainly consists of four blocks: pvariables, cpfs, reward and instance. The pvariables block declares state fluents and action fluents. The transitions of state fluents are described in cpfs block. The reward block defines the immediate reward, and the instance block defines the initial state.   

The RDDL description below illustrates an example of the Waterworld problem (partial). The state fluent \verb|ba_velo(ball, dir)| means the velocity of `ball' in direction `dir'. The action fluent \verb|ag_velo(dir)| controls the velocity of the agent in direction `dir'. The transition function \verb|ba_velo'(?b, ?d)| defines that in the next state how the value of \verb|ba_velo(?b, ?d)| will be changed depending on whether it hits the wall (\verb|crash(?b, ?d)|). The instance \verb|inst| sets up the initial value of \verb|ba_velo(?b, ?d)|.  The reward block will be introduced later.

\begin{small}
\begin{verbatim}
pvariables {
    ba_velo(ball, dir): 
    {state-fluent, real, default = 0.0};
    ag_velo(dir): 
    {action-fluent, real, default = 0.0};
    ...
};
cpfs{
    ba_velo'(?b, ?d) = 
        if(crash(?b, ?d)) then -ba_velo(?b, ?d) 
        else ba_velo(?b, ?d);
    ...
};
instance inst {
    init-state {
        ba-velo(b1, x) = -0.66; ba-velo(b1, y) = -1.90;
        ...
    };
    ...
}
\end{verbatim}
\end{small}

Intuitively speaking, the pvariables are encoded to the observation space and action space in Gym via the pyRDDLGym framework. For instance, the state fluent \verb|ba_velo| corresponds to a key-value pair in the observation space (Dict type)
\verb|{'ba_velo':Box(-inf, inf, (1,), float32)}|. The value in the pair is a data type of Box, representing a continuous space with one dimension, ranging from negative infinity to positive infinity. The action fluent \verb|ag_velo| corresponds to the action space \verb|{'ag_velo':Box(-inf, inf, (1,), float32)}| similarly. Both the transition functions in cpfs and the reward are simulated and calculated respectively in the step function of Gym. 
The instance block is wrapped in the reset function of Gym.   
\end{myexample}

\section{Exploiting DFA Structure in RL}
This section illustrates our approach in detail. As shown in Figure 2, in the first place, a non-Markovian reward specified by LTL$_f$ is encoded in an MDP (RDDL) by means of DFA. Then, the encoded MDP is converted to a Gym environment by pyRDDLGym. During the training process, we classify experiences based on DFA with respect to the degree of task completion.
\begin{figure}
\centering
\includegraphics[width=0.5\textwidth]{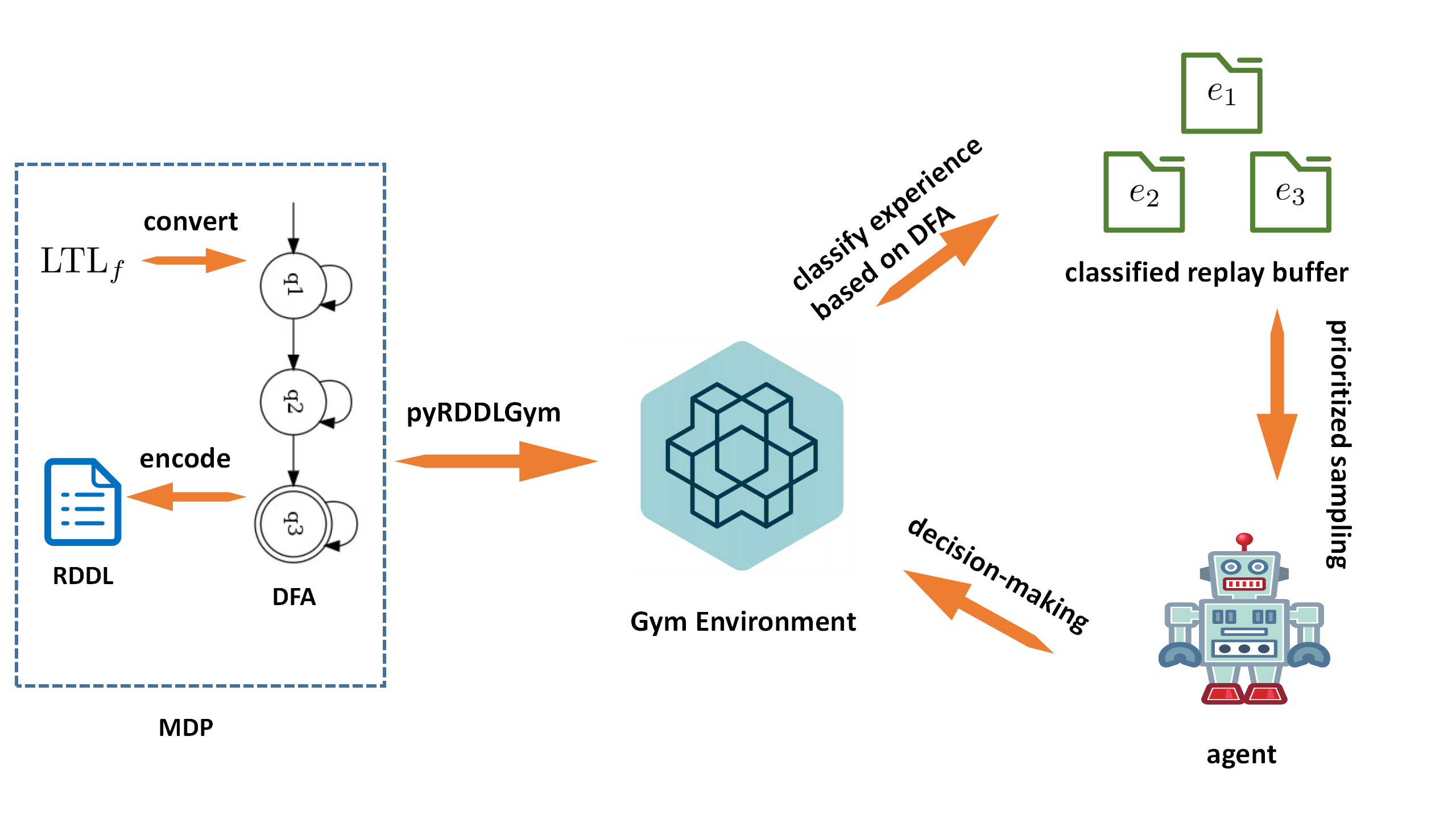}
\caption{Overview of our approach} 
\end{figure}
\subsection{Encoding NMRDP into MDP}
Given an NMRDP, we first encode it into a standard MDP. The encoding process comprises three steps. In what follows, we formalize each step together with exemplifying the instance of \emph{Waterworld} mentioned before.
\begin{itemize}
    \item \textbf{Step 1} A non-Markovian reward
in an NMRDP $NM$ is defined in terms of a tuple $\langle \varphi,r \rangle$, where $\varphi$ is an LTL$_f$ formula and $r$ is the associated reward.
    \item \textbf{Step 2} Transform $\varphi$ into a DFA $\mathcal{A}_\varphi=\langle Q,\Sigma,\delta,q_0,F\rangle$.
    \item \textbf{Step 3} Construct an MDP $M$ from $NM$ and $\mathcal{A}_\varphi$.
\end{itemize}

The first two steps are obvious. Consider Example 1, the non-Markovian reward can be defined as $\langle \varphi,100 \rangle$ in Step 1. Then $\varphi$ is transformed into a DFA in Step 2, as shown in Figure 3. Note that state $q2$ is an error state.

Given an $NM = \langle S, A, R, P, \gamma, s_{0} \rangle$, suppose the encoded $M$ is $\langle S', A, R', P', \gamma, s_0' \rangle$ at Step 3. We introduce an extra variable $f_Q$ of an enumerated type, which ranges over $Q$. $f_Q$ is used to denote the current activated state of $\mathcal{A}_\varphi$. Concretely, the encoding is formalized below.

\begin{itemize}
    \item States: $S'=\{s \cup \{f_Q=q\} \mid s \in S, q\in Q\}$
    \item Reward function: $R'(s_t',a_t,s_{t+1}')=$\\$\left\{\begin{matrix}  r \in \mathbb{R} ,& \text{if $f_Q = q_f, q_f\in F, a_t \in A$,}\\ 
    &\text{$f_Q=q_f \in s_{t+1}'$ and $s_t', s_{t+1}' \in S'$} \\  0,& \text{otherwise}\end{matrix}\right.$
    \item Transition probability: $P'(s_{t+1}'\mid s_t',a_t)=$\\$\left\{\begin{matrix} P(s_{t+1}\mid s_t,a_t), & \text{ if $\langle f_Q=q,\nu,f_Q=q' \rangle \in \delta$,} \\  & \text{$s_t \models \nu$, $\nu \in \Sigma$,}\\ &\text{ $s_{t}'=s_{t}\cup \{f_Q=q\}$ and }\\
    &\text{$s_{t+1}'=s_{t+1}\cup \{f_Q=q'\}$} \\  0,& \text{otherwise}\end{matrix}\right.$\\
    \item Intial state: $s_0'=s_0\cup \{f_Q = q_0\} $
\end{itemize}

 Each encoded state $s' \in S'$ combines the original state $s$ with $f_Q$. Obviously, the initial state $s_0'$ consists of $s_0$ and $f_Q = q_0$. The set of actions remains the same. In addition, $NM$ receives a reward $r$ when $f_Q = q_f$ ($q_f$ is an accepting state), which indicates $\varphi$ is satisfied. 
The transition probability $P'(s_{t+1}'\mid s_t',a_t)$ equals to $P(s_{t+1}\mid s_t,a_t)$ if there exists a possible DFA transition enabled from $s_t$ to $s_{t+1}$, and 0 otherwise. Practically, we apply the encoding method in the RDDL model. 

In \cite{miao2022approach,camacho2018non}, a set of propositional variables represents the set of DFA states, one variable for one state. Since there is only one activated state in DFA at each time step, $f_Q$ is sufficient for representing the dynamics of DFA transitions. Reducing the amount of state space at the time of RL training is beneficial.
Suppose the amount of original state space is $|S|$, and the state space in $M$ is linear in the size of $|S|$, i.e. $|S'|=|Q| \times |S|$. However, the increase of state space is exponential using the encoding of \cite{miao2022approach,camacho2018non}, i.e. $|S'|=2^{|Q|}  \times |S|$.
\begin{figure}
\centering
\includegraphics[width=0.35\textwidth]{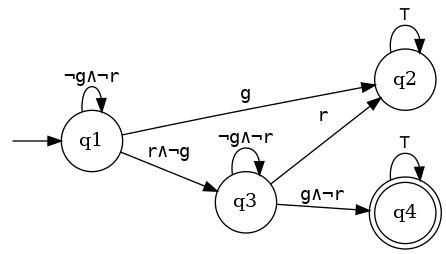}
\caption{DFA of $\varphi = \left ( \neg r \wedge \neg g \right ) \text{U}((r \wedge \neg g) \wedge  \bigcirc ((\neg r \wedge  \neg g)\text{U} (g \wedge \neg r))) $} \label{fig3}
\end{figure}

\begin{myexample}[DFA encoding] 
The following RDDL snippet describes the transitions of DFA in Figure 3, as well as the reward function and termination conditions. \verb|fQ| is defined as a state fluent, which is an enumerated variable with four values \verb|@q1|, \verb|@q2|, \verb|@q3| and \verb|@q4| corresponding to the four states in $Q$. In the cpfs block, each conditional statement corresponds to a transition, determining the successor value of \verb|fQ|. The reward will be received until \verb|fQ| equals to \verb|@q4|. 

In order to avoid lengthy trails, RDDL provides a termination block to truncate an episode. In our encoding, we put all the accepting and error states in the termination block, i.e. \verb|fQ=@q2| and \verb|fQ=@q4|. 
\begin{small}
\begin{verbatim}
pvariables {
    fQ : {state-fluent,{@q1, @q2, @q3, @q4},default=@q1};
};
cpfs{
    fQ' = if(fQ == @q1 ^ (~r ^ ~g)) then @q1
          else if(fQ == @q1 ^ (g ^ ~r)) then @q2
          else if(fQ == @q2 ^ true) then @q2
          else if(fQ == @q1 ^ r) then @q3
          else if(fQ == @q3 ^ (r ^ ~g)) then @q2
          else if(fQ == @q3 ^ (~r ^ ~g)) then @q3
          else if(fQ == @q3 ^ g) then @q4
          else if(fQ == @q4 ^ true) then @q4
          else fQ;
};
reward = 100*(fQ == @q4);
termination {fQ == @q2; fQ == @q4;};
\end{verbatim}
\end{small}
\end{myexample}

\subsection{Experience Classification based on DFA}  
In this subsection, we propose a method to improve RL algorithms' training speed and policy quality by prioritizing experiences. 

For improving the training speed of RL algorithms, most current research is devoted to improving the sampling probability of high-value experiences. Schaul et al. propose the technique of prioritized experience replay (PER) \cite{schaul2015prioritized}. They develop a framework for prioritizing experience for online algorithms, which allows agents to learn from transitions sampled with non-uniform probability proportional to their temporal-difference (TD) error. They set DQN as a baseline, significantly improving its performance in most Atari games with PER. Although PER outperforms in discrete action domains, current state-of-the-art RL algorithms with actor networks cannot effectively train in continuous action domains using PER when experiences with large TD errors are involved.

Hou et al. successfully combined PER with deep deterministic policy gradient (DDPG), which applied PER to continuous action domains \cite{hou2017novel}. The experimental results show that DDPG with prioritized experience replay can reduce the training time and improve the stability of the training process. Horgan et al. propose a distributed architecture relying on PER to focus only on the most significant data generated by the actors \cite{horgan2018distributed}. This architecture enables agents to learn effectively from orders of magnitude more data than previously possible and achieve state-of-the-art results in a wide range of discrete and continuous tasks.

Borrowing the idea from \cite{camacho2018non}, we prioritize DFA states automatically according to the automata-based potentials. On this basis, we classify experiences into different categories associated with the rank of DFA states. It should be noted that our method is suitable for a series of off-policy algorithms.  

Intuitively, the closer a DFA state to the accepting states, the higher the probability of accomplishing the task. Hence, we introduce criteria to prioritize DFA states, i.e., expected path length to accepting states.  

Given a DFA $\mathcal{A}=\langle Q,\Sigma,\delta,q_0,F \rangle$ and the number of categories $3 \le N \le |Q|$, Algorithm 1 prioritizes $Q$ with respect to the expected path length and returns the ranks of $Q$ ($0 \le rank(q) \le N-1, q \in Q$). We calculate the average length $L(q)$ of non-looping paths from each non-accepting state $q \in Q \setminus F$ to the accepting states (lines 2-3). The ranks of error states and accepting states are $N-2$ (if they exist) and $N-1$ respectively. The ranks of the other states are from 0 to $N-3$, calculated by the equation in line 6. $L_{max} / L_{min}$ is the maximum/minimum length among $\{L(q)\mid q\in Q \setminus F \setminus E\}$. The interval $L_{max} - L_{min}$ is equally divided into $N-2$ sub-intervals, corresponding to the ranks $\{0,...,N-3\}$. 

\begin{algorithm}
\caption{Automatic States Prioritization (ASP)}
\textbf{Input:} 
    $\mathcal{A} =\langle Q,\Sigma,\delta,q_0,F \rangle, N$ \\
\textbf{Output:} 
     $rank$ 
\begin{algorithmic}[1]
\State Initialize expected path length $L(q) \gets 0$, for all $q \in Q$ 
\For{$q \in Q \setminus F$}
    \State $L(q) \gets$ compute the expected path length of $q$
\EndFor
\For{$q \in Q$}
    \If{$q \in Q \setminus F \setminus E$}
        \State $rank(q) \gets \left\lfloor N-3-\frac{(L(q)-L_{min})(N-3)}{(L_{max}-L_{min} )} + \frac{1}{2}\right\rfloor$
    \ElsIf{$q \in E$}
        \State $rank(q) \gets N-2$
    \Else
        \State $rank(q) \gets N-1$
    \EndIf
\EndFor
\State \textbf{return} $rank$
\end{algorithmic}
\end{algorithm}

An experience $e$ is defined as $\langle s,a,s',r \rangle$, where the agent takes action $a$ in state $s$, enters the next state $s'$, and receives reward $r$. Based on the rank of DFA states, we classify experiences into $N$ categories, having different priorities. 

Algorithm 2 (referred to as EC) describes the training process including experience classification, and its input comprises an MDP $M$, a DFA $\mathcal{A}$, the exponent $\alpha$, the update interval of sampling probability $K$ and the number of categories $N$. Line 2 is to partition the storage space of the replay buffer $\mathcal{B}$ into $N$ separate parts.  

\begin{algorithm}
\caption{Experience Classification (EC)}
\textbf{Input:} 
    $M =\langle S, A, R, P, \gamma, s_{0}\rangle, \mathcal{A},\alpha, K, N$ 
\begin{algorithmic}[1]
\State $rank \gets$ ASP$(\mathcal{A}, N)$
\State $\{\mathcal{B}_0,...,\mathcal{B}_{N-1} \}\gets$ Divide replay buffer $\mathcal{B}$ into $N$ parts
\For{$l\gets 0$ \textbf{to} $num\_episodes$}
    \If{$l\bmod K=0$} 
        \State $\mathcal{P}(i) \gets  {\textstyle (|\mathcal{B}_i|p_i^{\alpha })}/ {\textstyle (\sum_{j=0}^{N-1}|\mathcal{B}_j| p_j^\alpha})$ \textbf{for} $i=0$ \textbf{to} $N-1$
    \EndIf
    \State Initialize $s\gets s_0$ 
    \While{$s$ is not terminal}
    \State Choose action $a\sim \pi(s), a \in A$
    \State Take action $a$ and observe $s'$ and $r$
    \State Reshape reward $r\gets r + \gamma\rho(q') - \rho(q), f_Q = q \in s$ and $f_Q = q' \in s'$
    \State Save experience $e=\langle s,a,r,s' \rangle $ in $\mathcal{B}_{rank (q')},  f_Q=q' \in s'$
    \State Sample experience batch from $\mathcal{B}$ based on $\mathcal{P}$ 
    \State Train and update policy $\pi$ 
    \State $s\gets s'$
    \EndWhile
\EndFor 
\end{algorithmic}
\end{algorithm}

We classify experiences based on $rank(q)$, where $f_Q=q$ is in the next state of an experience. The experiences in a category share one thing in common: their next states have the same $rank$ value. Here we adopt the classic experience sampling probability technique proposed in \cite{schaul2015prioritized}. As shown in Equation (1), the sampling probability $\mathcal{P}(i)$ of category $i$ is adjusted according to its proportion and priority. The priority of category $i$ is defined as $p_i =  \mathcal{C}/ (N-rank(q))$, where $f_Q=q $ is in the next state of an experience in category $i$, $rank(q)=i$, and $\mathcal{C}$ is a positive constant. $|\mathcal{B}_i|$ indicates the number of experiences in $\mathcal{B}_i$. 
The exponent $\alpha$ determines how much prioritization is
balanced, with $\alpha=0$ corresponding to the uniform case. In lines 4-5, the algorithm periodically updates the sampling probability of each category.
\begin{equation}
\text{$\mathcal{P}(i) = {\textstyle (|\mathcal{B}_i|p_i^{\alpha })}/ {\textstyle (\sum_{j=0}^{N-1}|\mathcal{B}_j| p_j^\alpha})$} 
\end{equation}

 Lines 7-14 describe the general process of policy updates. The idea of reward shaping is applied in line 10. Reward shaping is a common technique in RL whereby additional training rewards are used to guide the agent \cite{ng1999policy}. Let the encoded MDP $M$ be $\langle S, A, R, P, \gamma, s_0 \rangle$. Concretely, the reshaped reward function is shown as follows:
\begin{equation}
\begin{aligned}
\mathcal{R}(s_t, a_t, s_{t+1}) &= R(s_t,a_t,s_{t+1}) + \gamma\rho (q')-\rho(q)\nonumber
\end{aligned}
\end{equation}
where $s_t, s_{t+1} \in S$, $f_Q = q' \in s_{t+1}$, $f_Q = q \in s_t$, $R$ is the reward function of $M$, $\rho:Q\to \mathbb{R}$ is the potential function and $\gamma$ is the discount factor. Except for the error states, we define the potential of DFA states the same as the priority of the category they belong to, i.e. $\rho(q)=p_i$, where $rank(q)=i$ and $q \in Q \setminus E$. The potentials of the error states are set to $C/N$ (the lowest value among $Q$).
\begin{figure}
\centering
\includegraphics[width=0.475\textwidth]{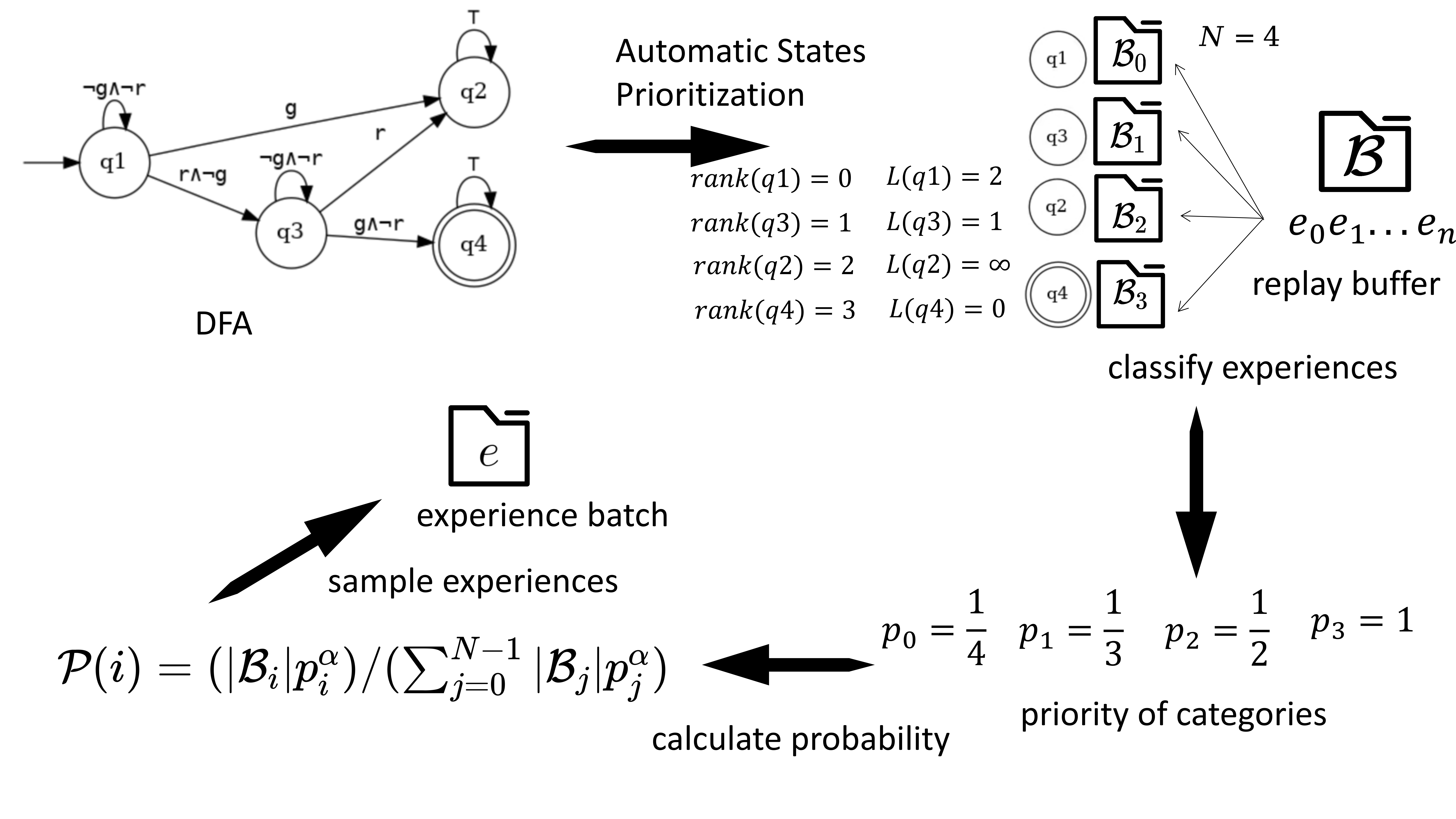}
\caption{An example of EC}
\end{figure}

\begin{myexample}[Experience classification]

  Figure 4 provides an example of EC. Given the DFA in Figure 3, we prioritize DFA states into four ranks which, sorted from low to high, are $q_1, q_3, q_2$ and $q_4$ in turn. Their ranks are $0,1,2,3$ respectively. The replay buffer $\mathcal{B}$ is divided into four parts, each of which corresponds to one rank. An experience is only added to one part of the replay buffer according to its associated DFA state. The priorities of the 4 experience categories are $p_0 = 1/4, p_1 =1/3, p_2 =1/2$ and $p_3=1$, arranged in ascending order ($\mathcal{C}=1$). Finally, a batch of experiences is sampled based on the probability Equation (1).
\end{myexample}

\begin{table*}
		\caption{Task descriptions of \emph{Waterworld} and \emph{Cartpole}}\
        \resizebox{\textwidth}{!}{
		\label{tbl1}
        \begin{tabular}{cc|c|c}
        \toprule
\emph{Waterworld} & \multicolumn{3}{c}{Informal/Formal descriptions} \\
\midrule
\multirow{2}{*}{1} & \multicolumn{3}{c}{red strict-then blue strict-then green} \\
& \multicolumn{3}{c}{$(\neg r \wedge \neg b \wedge \neg g) \text{U}((r \wedge \neg b \wedge \neg g) \wedge \bigcirc ((\neg r \wedge \neg b \wedge \neg g)\text{U}((b \wedge \neg r \wedge \neg g) \wedge \bigcirc ((\neg r \wedge \neg b \wedge \neg g) \text{U}(g \wedge \neg r \wedge \neg g) )) )) $}\\

\multirow{3}{*}{2} & \multicolumn{3}{c}{red strict-then blue strict-then green strict-then red strict-then green then blue} \\
& \multicolumn{3}{c}{$(\neg r \wedge \neg b \wedge \neg g) \text{U}((r \wedge \neg b \wedge \neg g) \wedge \bigcirc ((\neg r \wedge \neg b \wedge \neg g)\text{U}((b \wedge \neg r \wedge \neg g) \wedge \bigcirc ((\neg r \wedge \neg b \wedge \neg g) \text{U}$ }\\
&\multicolumn{3}{c}{ $((g \wedge \neg r \wedge \neg b) \wedge \bigcirc ((\neg r \wedge \neg b \wedge \neg g) \text{U}((r \wedge \neg b \wedge \neg g)\wedge \bigcirc((\neg r \wedge \neg b \wedge \neg g) \text{U} ((g \wedge \neg r \wedge \neg b) \wedge \bigcirc((\neg r \wedge \neg b \wedge \neg g) \text{U} (b \wedge \neg r \wedge \neg g)))) ) )) )) )) $}\\

\multirow{3}{*}{3} & \multicolumn{3}{c}{(red strict-then blue strict-then green) and (black strict-then white strict-then grey) }\\
& \multicolumn{3}{c}{$(\neg r \wedge \neg b \wedge \neg g) \text{U}((r \wedge \neg b \wedge \neg g) \wedge \bigcirc ((\neg r \wedge \neg b \wedge \neg g)\text{U}((b \wedge \neg r \wedge \neg g) \wedge \bigcirc ((\neg r \wedge \neg b \wedge \neg g) \text{U}(g \wedge \neg r \wedge \neg g) )) )) \wedge $ }\\
& \multicolumn{3}{c}{$(\neg B \wedge \neg W \wedge \neg G) \text{U}((B \wedge \neg W \wedge \neg G) \wedge \bigcirc ((\neg B \wedge \neg W \wedge \neg G)\text{U}((W \wedge \neg B \wedge \neg G) \wedge \bigcirc ((\neg B \wedge \neg W \wedge \neg G) \text{U}(G \wedge \neg B \wedge \neg W) )) )) $}\\
\midrule
\emph{Cartpole} & \multicolumn{3}{c}{Informal/Formal descriptions} \\
\midrule
\multirow{2}{*}{$4 \mid 5 \mid 6$} & move into $g_1$ then $g_2$ then $g_3$ & \text{move into $g_1$ then $g_2$ then $g_3$ then $g_4$ then $g_5$} & \text{move into $g_1$ then $g_2$ then $g_3$ then $g_4$ then $g_5$ then $g_6$ then $g_7$}\\
& $\Diamond (g_1 \wedge \bigcirc\Diamond(g_2 \wedge \bigcirc\Diamond(g_3)))$ & $\Diamond (g_1 \wedge \bigcirc\Diamond(g_2 \wedge \bigcirc\Diamond(g_3 \wedge \bigcirc\Diamond(g_4 \wedge \bigcirc\Diamond(g_5)))))$ & $\Diamond (g_1 \wedge \bigcirc\Diamond(g_2 \wedge \bigcirc\Diamond(g_3 \wedge \bigcirc\Diamond(g_4 \wedge \bigcirc\Diamond(g_5 \wedge \bigcirc\Diamond(g_6 \wedge \bigcirc\Diamond(g_7)))))))$\\
\bottomrule
        \end{tabular}
        }
	\end{table*}

\section{Experimental Evaluation}

In this section, we provide empirical evaluations of two typical problems: \emph{Waterworld} and \emph{Cartpole}, whose state spaces are continuous. Consider the \emph{Cartpole} problem \cite{tassa2018deepmind},
a pole is attached by an unactuated joint to a cart, which moves along a frictionless track. The pendulum is placed upright on the cart, and the goal is to balance the pole by applying forces in the left or right direction on the cart.

Our experiments are conducted on a MacBook Air with 8GB of RAM and an M1 chip running on macOS Ventura 13.1. We inherit RL algorithms provided by the Stable-Baselines3 (SB3) library to train the agent. SB3 is a set of reliable implementations of RL algorithms in PyTorch \cite{raffin2021stable}. The algorithms follow a consistent interface, making it simple to train and compare different RL algorithms. Among the off-policy algorithms of SB3, we choose TD3 (Twin Delayed DDPG) and SAC (Soft Actor-Critic) to perform the evaluations. TD3, a direct successor of DDPG, builds on Double Q-learning by taking the minimum value between a pair of critics to limit overestimation \cite{fujimoto2018addressing}. As the successor of Soft Q-Learning, SAC is an actor-critic deep RL algorithm based on the maximum entropy RL framework \cite{haarnoja2018soft}.

\begin{figure}
\centering
\includegraphics[width=0.3\textwidth]{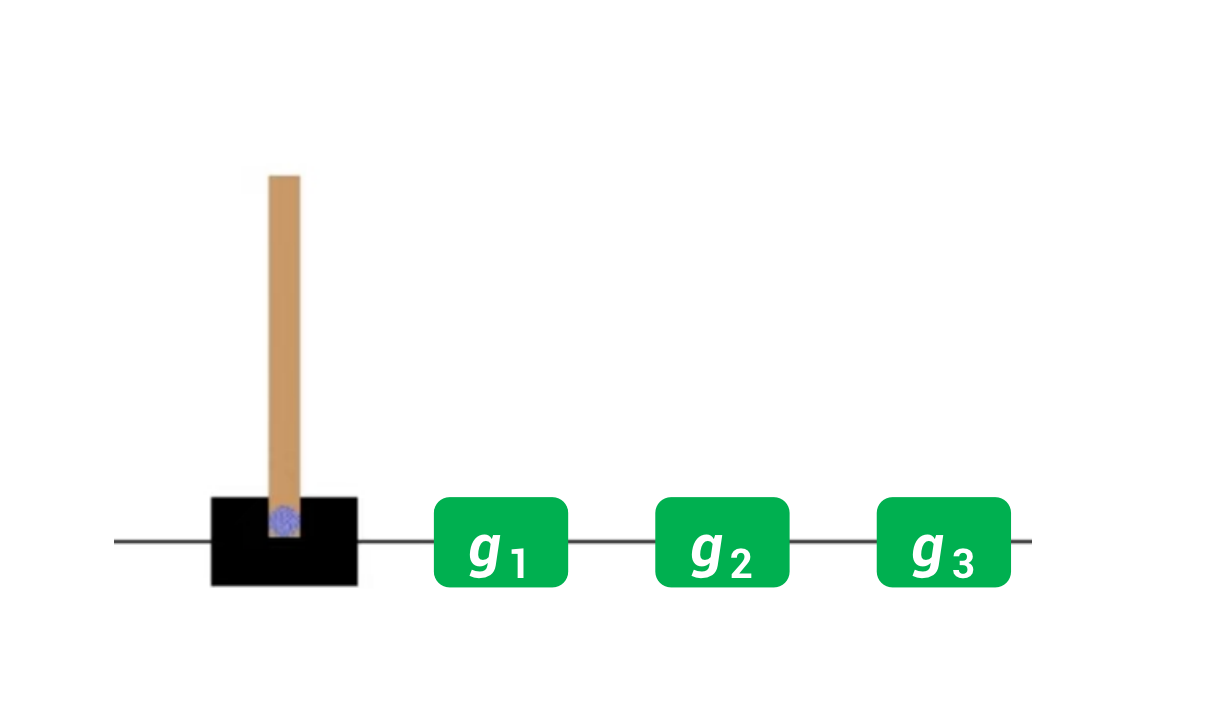}
\caption{A pictorial example of the \emph{Cartpole} problem} 
\end{figure}

The first experiment aims to evaluate the performance of EC.
Table 1 shows the non-Markovian tasks of the two problems, which are borrowed from \cite{icarte2022reward,hasanbeig2020deep} with slight changes. Task 1 is the same as the two most difficult tasks in \cite{icarte2022reward}. The difficulty of tasks 2 and 3 is far beyond task 1. Inspired by \cite{hasanbeig2020deep}, the track is divided into disjoint regions labeled by $g_1$, $g_2$, ..., as shown in Figure 5. Since the original tasks of \emph{Cartpole} \cite{hasanbeig2020deep}, specified in LTL, have infinite behaviors, we design new tasks 4 to 6, suitable for LTL$_f$. As an example, we illustrate tasks 1 and 4 to explain the meanings of the tasks. Task 1 requires the agent to touch red, green and blue balls in strict order. The goal of task 4 is to let the agent move into the colored regions $g_1$, $g_2$ and $g_3$ sequentially while maintaining the pole balance. The major difference between tasks 1 and 4 is whether the agent must finish the subgoals in strict order (strict-then/then). We employ the open source tool LTL$_f$2DFA to map LTL$_f$ formulae into DFAs \cite{fuggitti-ltlf2dfa}.

\begin{figure*}[h] 
		\centering 
		\includegraphics[width=\linewidth]{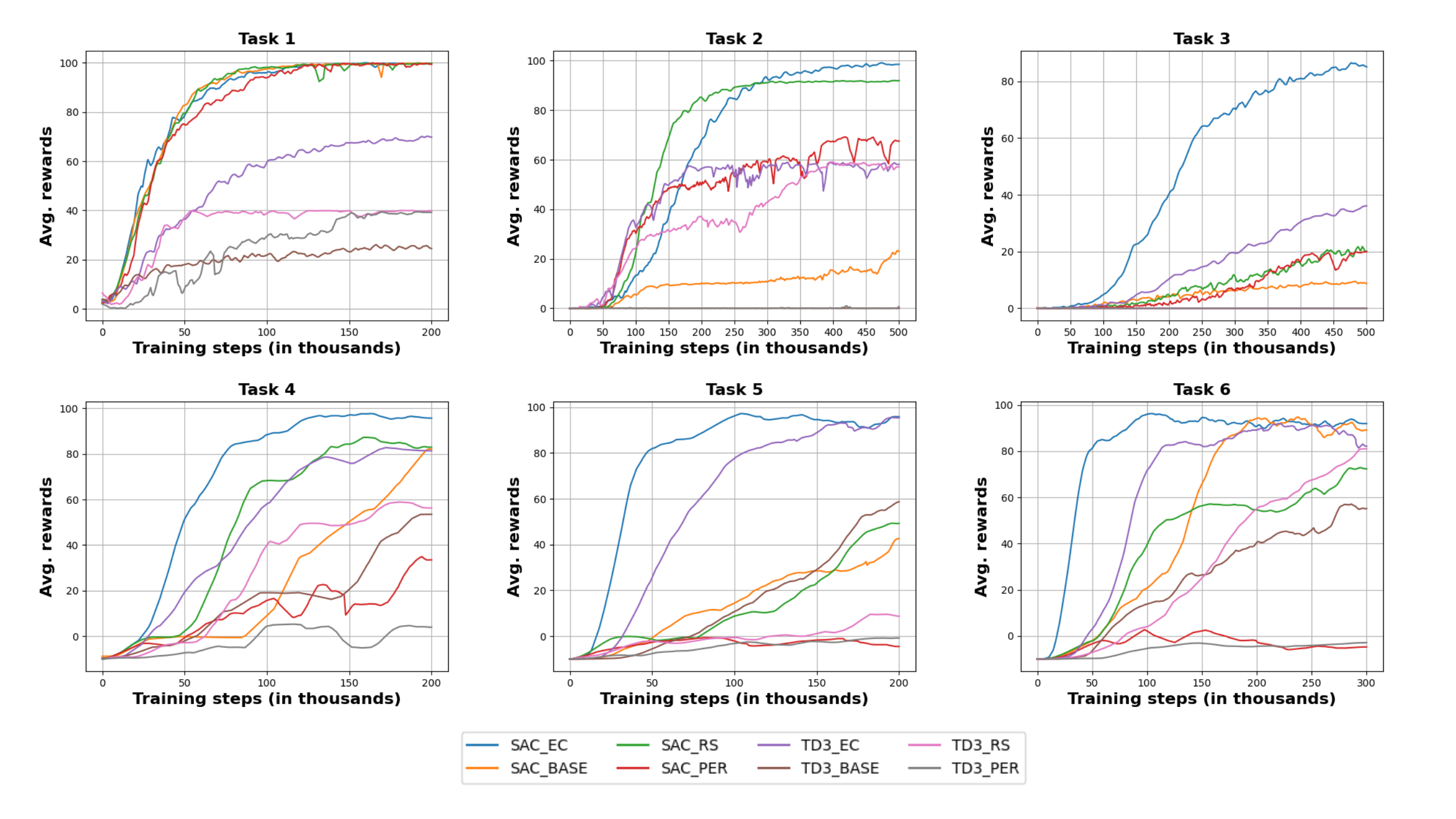}
		\caption{Results on the training speed} 
		\label{FIG:1} 
\end{figure*}

\begin{figure*}[h] 
		\centering 
		\includegraphics[width=\linewidth]{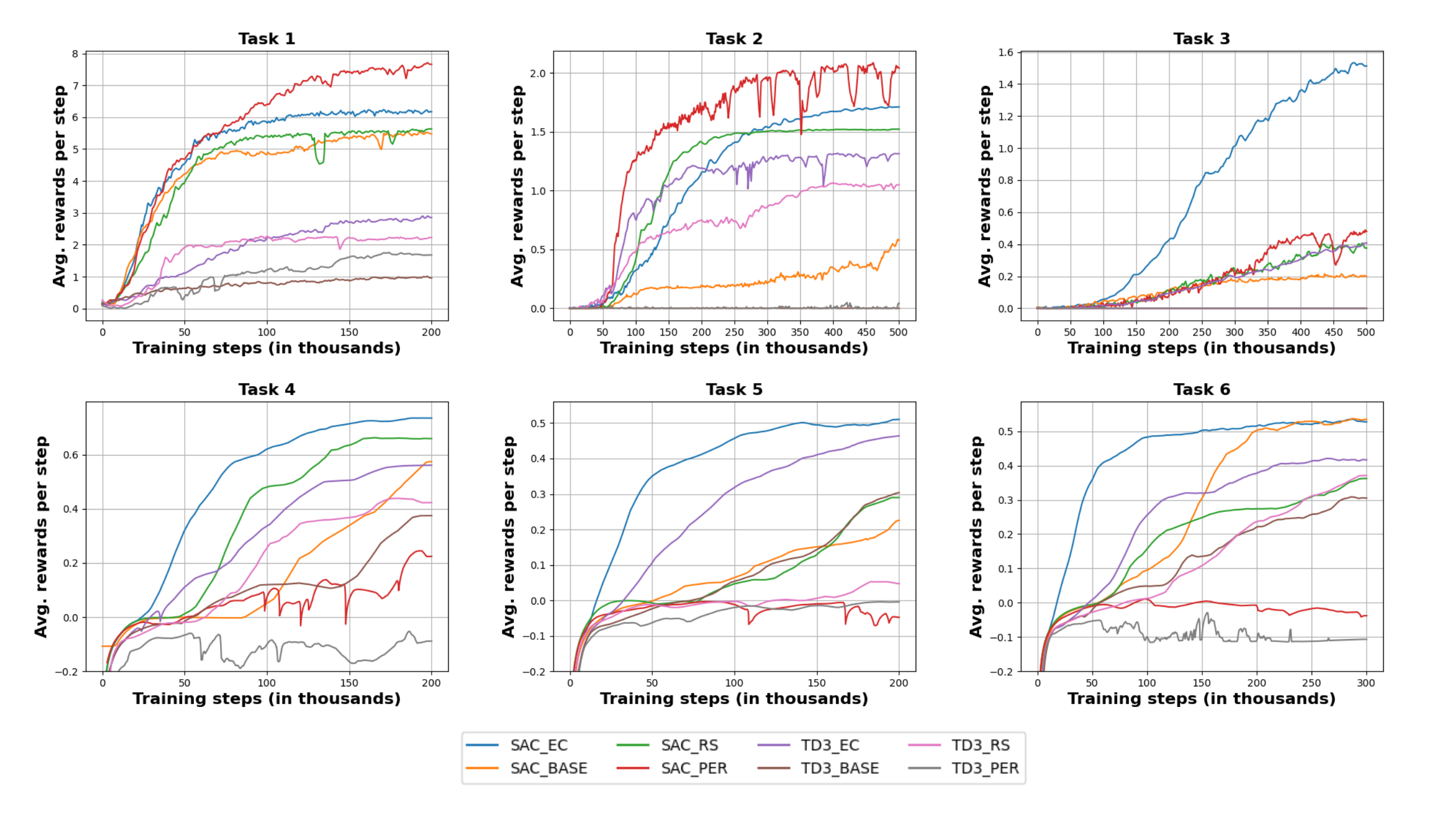}
		\caption{Results on the policy quality} 
		\label{FIG:1} 
\end{figure*}

 We randomly generate 10 \emph{Waterworld} maps with different positions and speeds of balls for each task. The boundary of each map is $20 \times 20$ or $30 \times 30$, including 9 or 18 balls inside. The pole of \emph{Cartpole} problem starts upright with 0 $rads$ and is located in the middle of the track. The width of the track is 7$m$, and that of a colored region is 1$m$. The length of the pole is 0.5$m$. The radius of a ball is 0.5. The speed of a ball ranges from -2 to 2 and that of the agent from -3.5 to 3.5. An episode terminates if the pendulum deviates more than 0.21$rads$ from the vertical position or the center of the cart reaches the edge of the track. The agent will receive a reward of 100 if it completes the task. A penalty -10 will be given when an episode of \emph{Cartpole} terminates, and no penalty is given at any step of \emph{Waterworld}. 

 We set exponent $\alpha=0.75$ in EC, and the steps of learning start as 1000 and the learning rate as $3e-4$. A Gaussian action noise with zero mean and a variance of 0.1 is added to TD3, which can help for hard exploration problems. The remaining parameters in SAC and TD3 are configured by default. We train ten instances using SAC and TD3 for each task and obtain the average performance.

As a brief summary, the experimental results show that EC can improve the training speed and policy quality of off-policy algorithms. Figure 6 shows the comparison of EC with other methods on the training speed.
 The horizontal axis represents the number of training steps, and the vertical axis represents the average rewards per episode. Here RS denotes the automatic reward shaping in \cite{icarte2022reward}, PER refers to the technique of prioritized experience replay in \cite{schaul2015prioritized}, and BASE represents the basic algorithms. Generally speaking, EC outperforms other methods. The only exception is Task 1 with SAC as the baseline, their reward curves are basically in coincidence. The reason is that task 1 is relatively simple and easy to solve for SAC itself.  Compared with EC in all tasks, the training speed of RS is slower than that of EC (even though RS can increase sparse rewards and guide training, making it better than BASE). Especially in more complex tasks, the performance gap between the two is very significant. Similarly, the training speed of PER is also slower than that of EC. Only in tasks 2 with SAC as the baseline, the training speed of PER achieves a much better performance than BASE. But in all other cases, the improvement effect of PER on BASE is not significant, and it may even substantially decreases the performance of BASE. The reason is that PER will prioritizing experiences with high TD error, and actor networks in SAC and TD3 cannot be effectively trained with transitions that have large TD errors. 
 
 Figure 7 shows the experimental results on the policy quality. The higher policy quality, the more accumulated rewards and the shorter length of success episodes. Here, the policy quality is reflected by the average rewards per step (vertical axis). It is obvious that the result is basically consistent with the training speed. All curves of EC surpass the other methods, except for tasks 1 and 2 with SAC where it falls behind PER. Table 2 shows the improvement of EC, RS and PER on the  policy quality.
  For instance, 154.54\% means that on the average, the rewards per step of EC is 154.54\% higher than that of BASE. Since TD3\_BASE receives zero rewards in tasks 2 and 3, we do not display the data, denoted by `-'. In some cases, SAC\_EC and TD3\_EC are capable of improving policy quality up to 499.53\% and 300.93\% respectively. Although PER has the highest improvement effect on SAC in tasks 1 and 2, it significantly reduces the policy quality of BASE in tasks 4 to 6.

  \begin{table}[]
    \centering
    \caption{Improvement of policy quality: EC, RS, PER vs. BASE}
    \resizebox{0.5\textwidth}{!}{%
    \begin{tabular}{cccccccc}
    \toprule
        \multicolumn{2}{c}{}&Task1& Task2 & Task3 & Task4& Task5 & Task6 \\
    \midrule
         \multirow{2}{*}{SAC}&EC & 15.05\% & 415.30\% & \textbf{499.53}\% & \textbf{206.34}\% & \textbf{466.92}\% & \textbf{58.24}\%  \\
         &RS & 1.34\% & 428.30\% & 40.91\% & 122.88\% & 8.46\% & -31.00\%  \\
         &PER & \textbf{29.13}\% & \textbf{608.25}\% & 60.24\% & -62.99\% & -143.25\% & -107.09\%  \\
         \multirow{2}{*}{TD3}&EC & \textbf{154.54}\% & - & - & \textbf{196.78}\% & \textbf{300.93}\% & \textbf{109.31}\%  \\
         &RS & 140.45\% & - & - & 95.76\% & -113.07\% & 3.85\%  \\
         &PER & 40.06\% & - & - & -223.82\% & -175.17\% & -180.00\%  \\
    \bottomrule
    \end{tabular}
    }
\end{table}

\begin{table}[]
    \centering
    \caption{Comparison in training time using optimized encoding}
    \resizebox{0.5\textwidth}{!}{%
    \begin{tabular}{cccccccc}
    \toprule
        \multicolumn{2}{c}{}&Task1& Task2 & Task3 & Task4& Task5 & Task6 \\
    \midrule
        \multicolumn{2}{c}{No. of DFA states} & 5& 8 & 17 & 4 & 6 & 8 \\
         \multicolumn{2}{c}{Training steps (k)} & 200& 500 & 500 & 200 & 200 & 300 \\
         \multirow{2}{*}{SAC}& BASE& \textbf{2509} & \textbf{7240} & \textbf{11886} & \textbf{1770} & \textbf{1818} & \textbf{2882}\\ 
        &Camacho et al.&2876&8762 &13258&1930&1938&3114\\
         \multirow{2}{*}{TD3}& BASE & \textbf{2618} & 
         \textbf{8619} & \textbf{10395} & \textbf{1432} & \textbf{1566} & \textbf{2389}  \\
         &Camacho et al.&2947 & 9544 & 12624 & 1562 & 1687 & 2735 \\
    \bottomrule
    \end{tabular}
    }
\end{table}

To demonstrate the effectiveness of our optimized encoding method, we carry out a second experiment. The fluent encoding of the tasks above are additionally modeled using the idea of \cite{camacho2018non}, then trained by SAC and TD3 without EC. Table 3 reports the statistics of total training time (in seconds). The number of DFA states is closely related to the scale of the state space. Hence we list the number of DFA states in the first row.
In short, our optimized encoding method significantly reduce the training time for all tasks. Compared with \cite{camacho2018non}, our encoding reduces training time by an average of 12.23\% (12.67\% for TD3 and 11.80\% for SAC). In addition, the more the number of DFA states, the longer the training time, and most importantly, the more time saved by our encoding. For example, the training time of task 1 is about 2500s (5 DFA states), and that of task 3 is over 10,000s (17 DFA states). Correspondingly, in every ten thousand steps, the saved time of task 1 is only about 17s and that of task 3 is over 44s at most.

Finally, to demonstrate the impact of the balanced prioritization in EC, we design the the third experiment. We plot the speed curves of SAC\_EC in training tasks 3 and 6 with different values of the exponent, as shown in Figure 7. $\alpha=0$ indicates that experiences are sampled randomly during training. The higher the exponent, the higher the high-priority experiences being sampled. In the figure, all curves with $\alpha>0$ surpass that with $\alpha=0$, which validates that prioritizing experiences can improve the training speed. On average, the method improves the training speed by xxx \%, with the highest increase of xxx\% when $\alpha=0.25$.

\section{Summary and Discussion}
 In this paper, we propose a novel RL approach to train non-Markovian tasks specified by temporal logic LTL$_f$. More specifically, an encoding method is introduced to transform a non-Markovian task (NMRDP) into a Markovian task (MDP), and an approach using prioritized experience replay technique based on DFA structures is proposed to improve the training speed and policy quality. We emphasize that LTL$_f$ does not bind to our approach. Other temporal logics may also be applicable.
 The experimental results on benchmark problems demonstrate the effectiveness and feasibility of our approach. In future work, we plan to further exploit the DFA structures as in \cite{hasanbeig2020deep} to break down complex tasks into simple composable sub-tasks or modules. We believe that there is significant potential in parallel training of such sub-tasks in conjunction with our current approach.

	\bibliographystyle{model5-names}





\end{sloppypar}
\end{document}